\documentclass[sigconf]{aamas} 

\usepackage{balance} 

\usepackage{algorithm}
\usepackage{algorithmicx}
\usepackage{algpseudocode}
\usepackage{amsmath}
\usepackage{amsthm}
\usepackage{array}
\usepackage{graphbox}
\usepackage{graphicx}
\usepackage{setspace}

\graphicspath{{anc/}}

\setcopyright{ifaamas}
\acmConference[AAMAS '21]{Proc.\@ of the 20th International Conference on Autonomous Agents and Multiagent Systems (AAMAS 2021)}{May 3--7, 2021}{Online}{U.~Endriss, A.~Now\'{e}, F.~Dignum, A.~Lomuscio (eds.)}
\copyrightyear{2021}
\acmYear{2021}
\acmDOI{}
\acmPrice{}
\acmISBN{}

\acmSubmissionID{355}

\title[Stratified Experience Replay]{Stratified Experience Replay: Correcting Multiplicity Bias in Off-Policy Reinforcement Learning}
\subtitle{Extended Abstract}

\author{Brett Daley}
\affiliation{
  \institution{Northeastern University}
  \city{Boston}
  \state{MA}
  \country{USA}}
\email{b.daley@northeastern.edu}

\author{Cameron Hickert}
\affiliation{
  \institution{Harvard University}
  \city{Cambridge}
  \state{MA}
  \country{USA}}
\email{cameron_hickert@hks.harvard.edu}

\author{Christopher Amato}
\affiliation{
  \institution{Northeastern University}
  \city{Boston}
  \state{MA}
  \country{USA}}
\email{c.amato@northeastern.edu}


\newcommand{\BibTeX}{\rm B\kern-.05em{\sc i\kern-.025em b}\kern-.08em\TeX}

\makeatletter
\renewcommand{\ALG@name}{Data Structure}
\makeatother

\newcommand{\tightsection}[1]{
    \vspace{-0.04in}
    \section{#1}
}

\begin{document}


\pagestyle{fancy}
\fancyhead{}


\maketitle 


\tightsection{Introduction}

Deep Reinforcement Learning (RL) methods rely on experience replay~\cite{lin1992self} to approximate the minibatched supervised learning setting;
however, unlike supervised learning where access to lots of training data is crucial to generalization, replay-based deep RL appears to struggle in the presence of extraneous data.
Recent works have shown that the performance of Deep Q-Network (DQN)~\cite{mnih2015human} degrades when its replay memory becomes too large~\cite{zhang2017deeper, liu2018effects, fedus2020revisiting}.

\begin{figure}[t]
    \centerline{\includegraphics[width=0.5\textwidth]{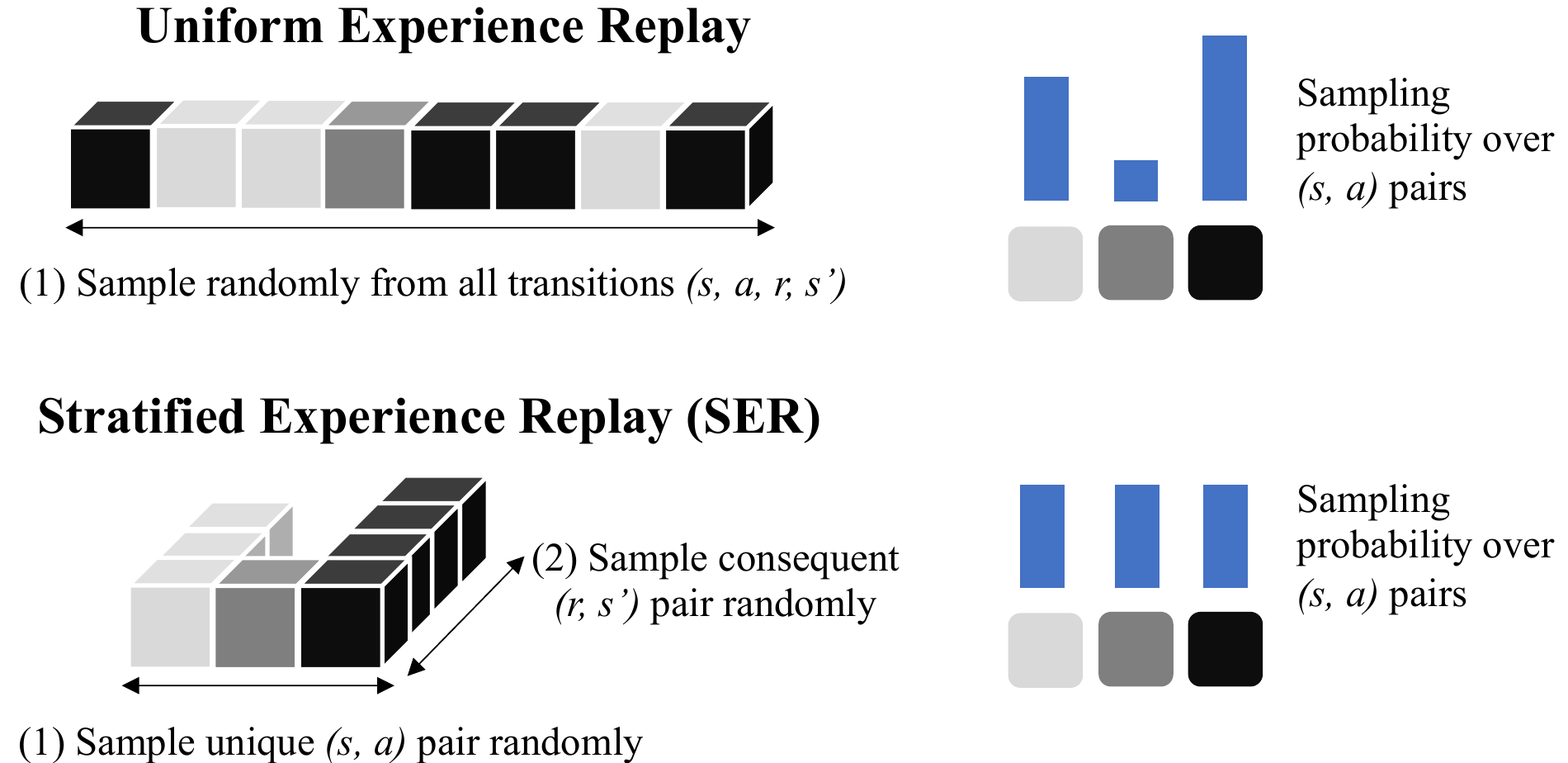}}
    \caption{A graphical comparison of uniform (top) and stratified (bottom) sampling strategies.}
    \label{fig:stratified_experience_replay}
\end{figure}

This suggests that outdated experiences somehow impact the performance of deep RL, which should not be the case for off-policy methods like DQN.
Consequently, we re-examine the motivation for sampling \emph{uniformly} over a replay memory, and find that it may be flawed when using function approximation.
We show that---despite conventional wisdom---sampling from the uniform distribution does not yield uncorrelated training samples and therefore biases gradients during training.
Our theory prescribes a special non-uniform distribution to cancel this effect, and we propose a stratified sampling scheme to efficiently implement it (see Figure~\ref{fig:stratified_experience_replay}).


\tightsection{Motivation}
\label{sect:motivation}

We begin by showing how bias arises under experience replay with function approximation by comparing Q-Learning~\cite{watkins1989learning} with its deep analog, DQN~\cite{mnih2015human}.
We model the environment as a Markov Decision Process (MDP) of the standard form
$(\mathcal{S}, \mathcal{A}, T, R)$~\cite{sutton2018reinforcement}.

Upon taking action $a \in \mathcal{A}$ in state $s \in \mathcal{S}$ and observing the resulting state $s' \in \mathcal{S}$, Q-Learning conducts an update on an entry $Q(s,a)$ of its lookup table.
Define the temporal-difference error as
$\delta(s,a,s') = R(s,a,s') + \gamma \max_{a' \in \mathcal{A}} Q(s',a') - Q(s,a)$ with discount factor $\gamma \in [0,1]$.
Since this particular error has probability $T(s,a,s') = \Pr(s' \mid s,a)$ of occurring, the \emph{expected} Q-Learning update can be computed:
\begin{align}
    \label{eq:qlearning_expected}
    Q(s,a) \gets Q(s,a) + \alpha \sum_{s' \in \mathcal{S}} \Pr(s' \mid s,a) \delta(s,a,s')
\end{align}

\begin{algorithm}[t]
    \caption{Stratified Replay Memory}
    \label{algo:stratified_replay_mem}
    \begin{algorithmic}
        \State {\bfseries Initialize} array $D$ of size $N$, hash table $H$, integer $i=0$
        \vspace{0.05in}
        \Procedure{insert}{$s$, $a$, $r$, $s'$}
            \If{$D$ is full}
                \State Get transition $(s_i, a_i, r_i, s'_i)$ from $D[i]$
                \State Pop queue $H[(s_i,a_i)]$;\ \ if now empty, delete key $(s_i,a_i)$
            \EndIf
            \State If $(s,a) \not\in H$, then $H[(s,a)] \gets empty\ queue$
            \State Push $i$ onto queue $H[(s,a)]$
            \State $D[i] \gets (s,a,r,s')$;\ \ $i \gets (i+1) \!\! \mod N$
        \EndProcedure
        \vspace{0.05in}
        \Function{sample}{\ }
            \State Sample state-action pair $(s,a)$ uniformly from the keys of $H$
            \State Sample integer $j$ uniformly from queue $H[(s,a)]$
            \State \Return transition $(s_j, a_j, r_j, s'_j)$ from $D[j]$
        \EndFunction
    \end{algorithmic}
\end{algorithm}


\begin{figure}[t]
    \centerline{
        \hfill
        \includegraphics[width=0.5\columnwidth]{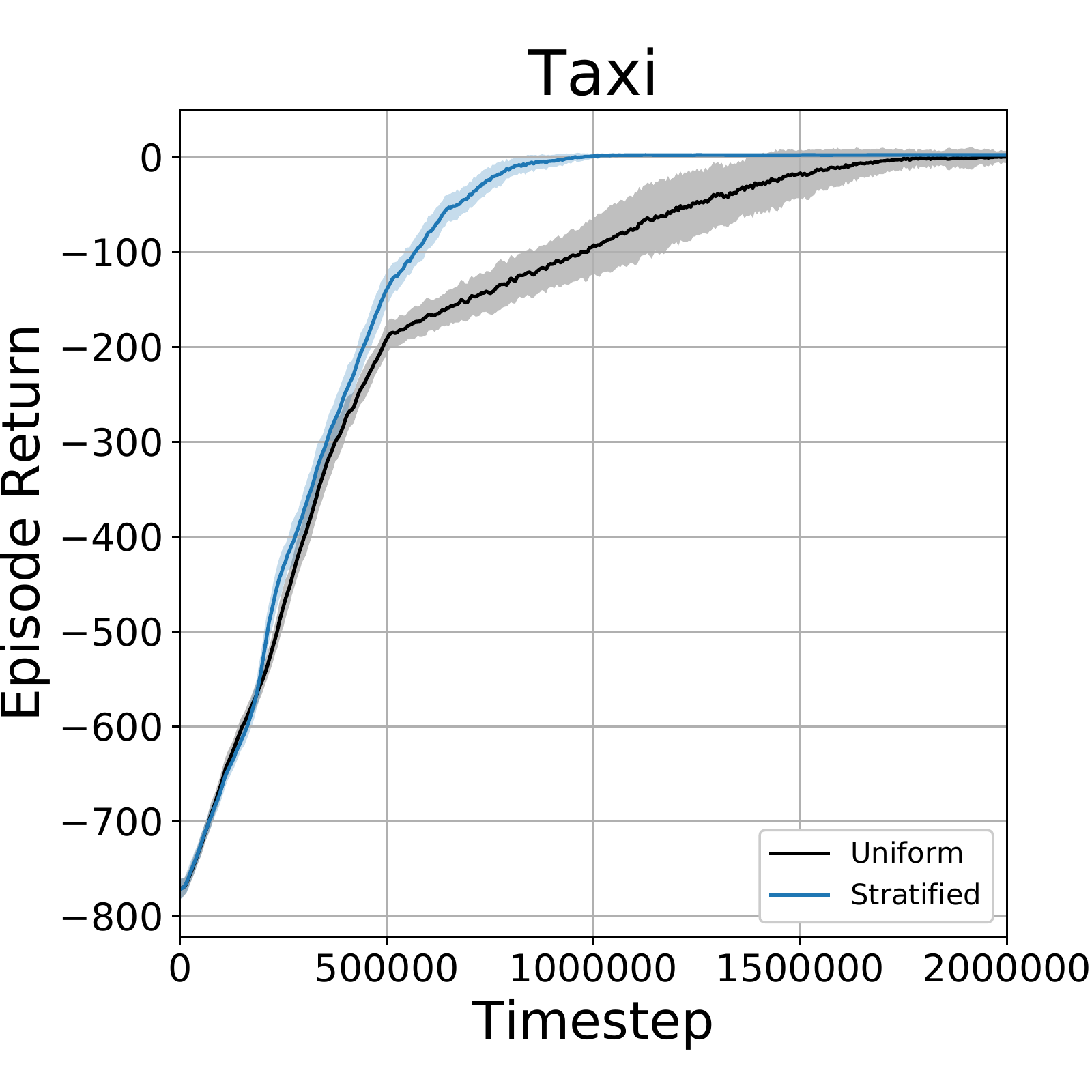}
        \hfill
        \includegraphics[width=0.5\columnwidth]{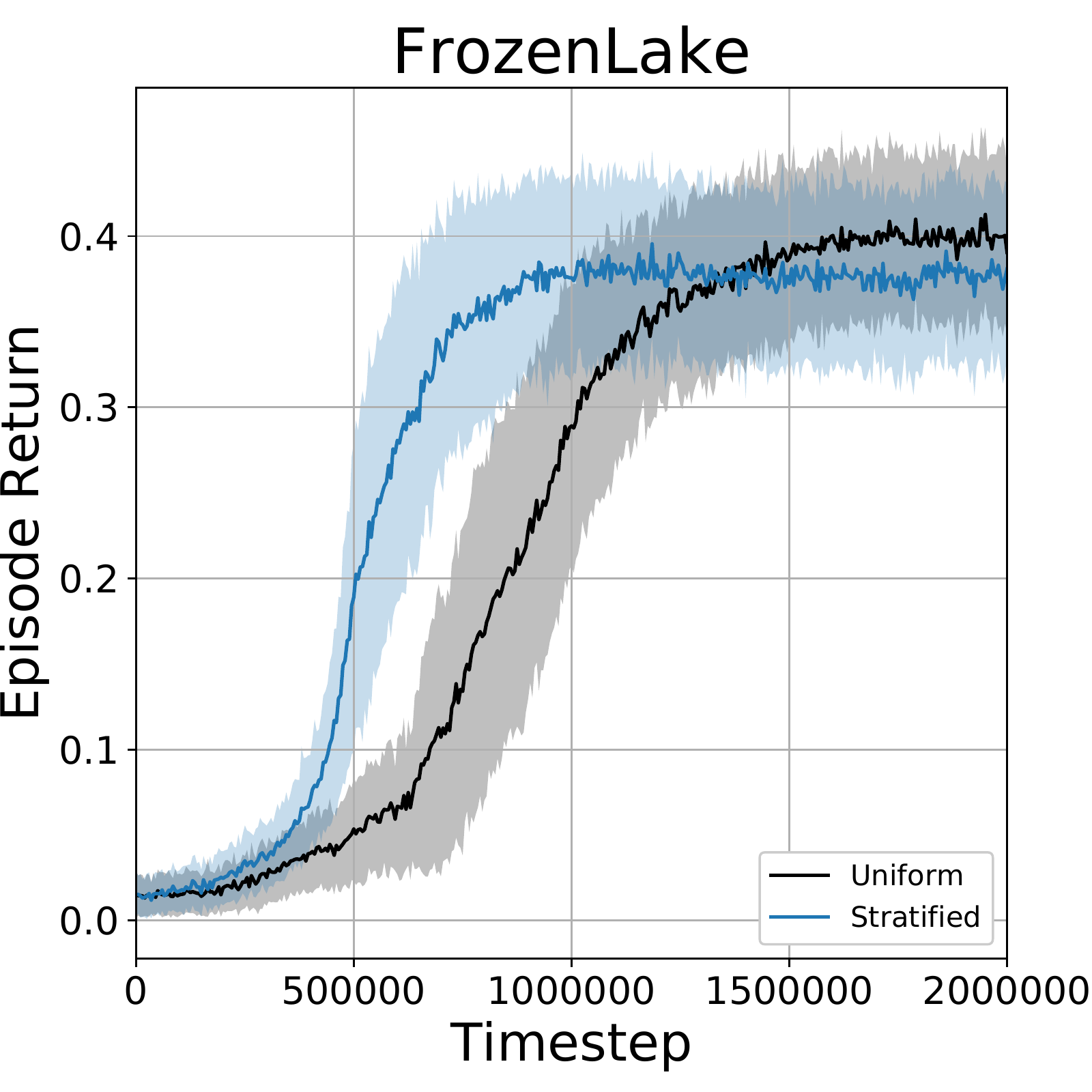}
        \hfill
    }
    \caption{
        SER performance compared against a uniform baseline on two environments, averaged over 100 trials.
    }
    \label{fig:toytext_experiments}
    \vspace{-0.2in}
\end{figure}

\noindent
where $\alpha \in [0,1]$ is the learning rate.
Importantly, the expected Q-Learning update is independent of the visitation frequency of the state-action pair $(s,a)$ as long as its probability of occurrence is nonzero.

Contrast this with DQN, which replaces the tabular lookup $Q(s,a)$ with a parametric function $Q(s,a;\theta)$ that is trained via stochastic gradient descent over a dataset of past experiences.
To facilitate our analysis, consider the theoretical case where DQN's replay memory $D$ has unlimited capacity and the agent executes a fixed behavior policy $\mu$ for an infinite duration before training.
We can deduce that a sample drawn uniformly from $D$ will have probability
$\Pr(s,a,r,s') = \Pr(s' \mid s,a) \Pr(s,a)$.\footnote{
    The reward $r = R(s,a,s')$ is deterministic and does not influence the probability.
}
Define the temporal-difference error as
$\delta(s,a,s') = R(s,a,s') + \gamma \max_{a' \in \mathcal{A}} Q(s',a';\theta^-) - Q(s,a;\theta)$ where $\theta^-$ is a time-delayed copy of $\theta$ that helps stabilize training.
The \emph{expected} DQN update can likewise be computed:
\begin{equation}
    \label{eq:dqn_expected}
    \theta \gets \theta + \alpha
    \Pr(s,a)
    \sum_{s' \in \mathcal{S}} \Pr(s' \mid s,a) \ \delta(s,a,s') \ \nabla_\theta Q(s,a;\theta)
\end{equation}
Note that this is analogous to (\ref{eq:qlearning_expected}) up to an additional factor of $\Pr(s,a)$.
This factor effectively scales the learning rate in proportion to how frequently the state-action pair occurs in the MDP under the policy $\mu$.
Hence, even under these rather favorable conditions (an unchanging policy with infinite training samples), DQN suffers from \emph{multiplicity bias} due to the uniform distribution.
Significantly, this bias is not unique to DQN and affects other off-policy deep RL methods like
DDPG~\cite{lillicrap2015continuous}, ACER~\cite{wang2016sample}, TD3~\cite{fujimoto2018addressing}, and SAC~\cite{haarnoja2018soft}.



\tightsection{Stratified Experience Replay}

According to our theory, an ideal experience replay strategy would sample state-action pairs in inverse proportion to their relative frequencies under the stationary distribution.
While it is not tractable to directly compute this distribution for high-dimensional environments, our agent has the advantage of a large replay memory at its disposal;
hence, sample-based approximations are feasible.

Recall from Section~\ref{sect:motivation} that the sampling probability under the uniform distribution factors:
$\Pr(s,a,r,s') = \Pr(s' \mid s,a) \Pr(s,a)$.
Dividing this by $\Pr(s,a)$ to eliminate the multiplicity bias, and then normalizing to make the probabilities sum to 1 over the set $\mathcal{S} \times \mathcal{A} \times \mathcal{S}$,
we arrive at the ideal sampling distribution:
${ \Pr(s' \mid s,a)\ /\ |\mathcal{S} \times \mathcal{A}| }$.
Remarkably, this indicates that we can sample from two uniform distributions in succession to counter multiplicity bias.
We call this \emph{Stratified} Experience Replay\footnote{
    Our approach should not be confused with the recent method of the same name~\cite{sharma2020stratified}.
} (SER) in which we first uniformly sample an antecedent state-action pair
$(s,a)$ from $D$ and then uniformly sample a consequent reward-state pair $(r,s')$ from the transitions observed in $(s,a)$.
By utilizing this two-step sampling strategy, we are able to achieve a reasonable approximation\footnote{
    It is not exact since, in practice, the replay memory will generally not contain $\mathcal{S} \times \mathcal{A}$ fully, nor will the experiences be collected from a single policy $\mu$.
    Future work that re-examines these simplifications could potentially improve empirical performance.
}
to the ideal distribution without needing to explicitly compute these probabilities.
Data Structure~\ref{algo:stratified_replay_mem} outlines an efficient implementation of SER that avoids an expensive search over the replay memory and thereby maintains a sampling cost of $O(1)$.

\begin{figure}[t]
    \centerline{
        \includegraphics[width=\columnwidth, height=0.5\columnwidth]{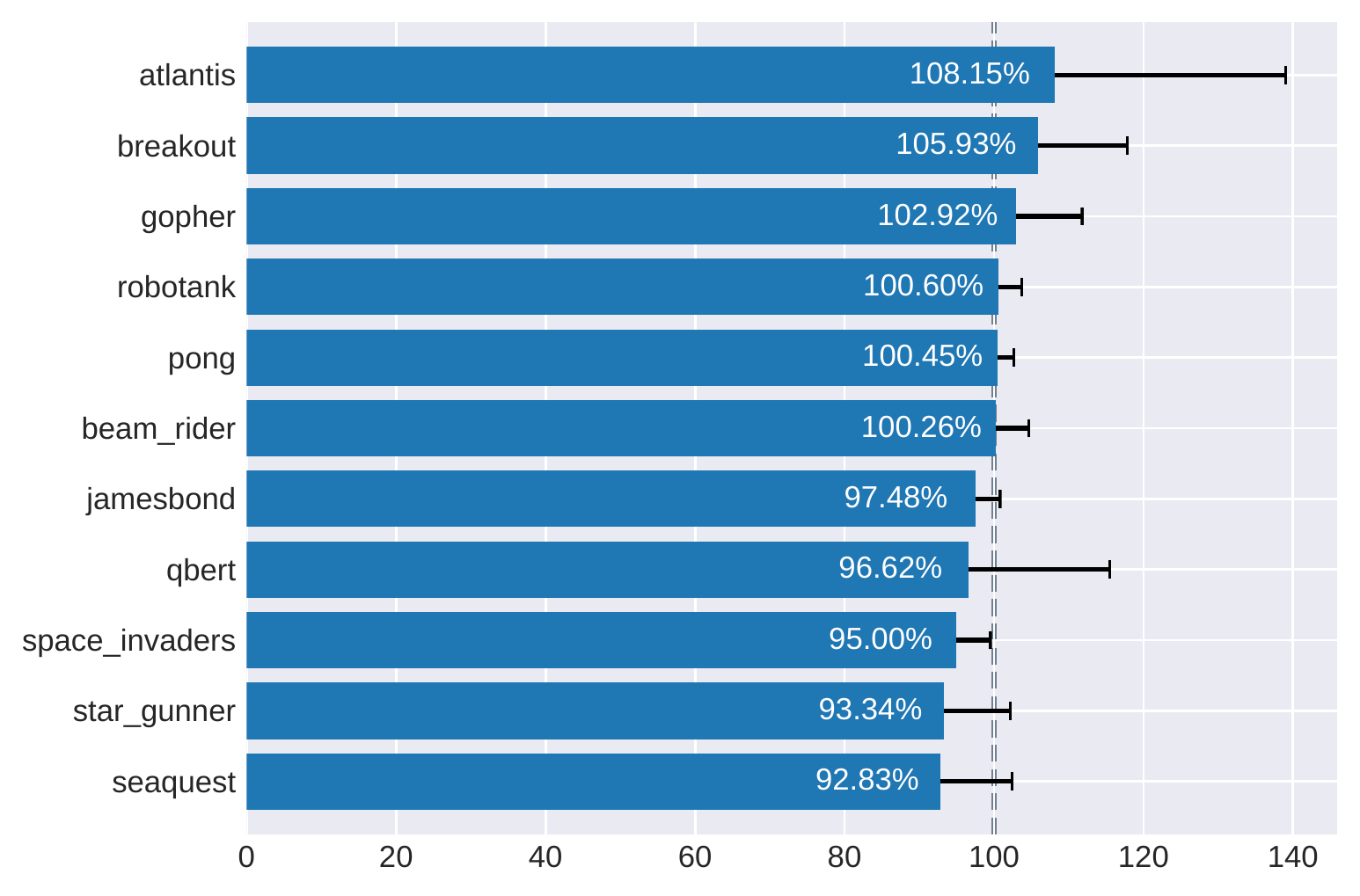}
    }
    \caption{
        Average episode score of SER throughout training on 11 Atari games, relative to that of the uniform baseline,
        ${ \text{i.e.}\ 100 \times (\text{stratified} - \text{random})\ /\ (\text{uniform} - \text{random}) }$.
    }
    \label{fig:atari_experiments}
\end{figure}


\section{Experiments}

Code and implementation details for all experiments are online.\footnote{
    \urlstyle{tt}
    \url{https://github.com/brett-daley/stratified-experience-replay}
}
All networks were optimized using Adam~~\cite{kingma2014adam}.
In our first experiment, we trained a two-layer tanh DQN to solve Taxi~\cite{dietterich2000hierarchical} and FrozenLake~\cite{brockman2016openai}, comparing the performance of SER against uniform experience replay.
SER helps the agent learn significantly faster just by changing the sampling distribution (Figure~\ref{fig:toytext_experiments}).

Our second experiment compared the two sampling strategies when training a convolutional DQN on 11 Atari 2600 games within the ALE~\cite{bellemare2013arcade} following the procedures in~\cite{mnih2015human} (excepting the use of Adam).
While SER improved average performance in a majority of the games (Figure~\ref{fig:atari_experiments}), the benefits were relatively modest compared to those of our first experiment.
This is likely due to the high-dimensional nature of the games, wherein the majority of state-action pairs are visited no more than once.

Nevertheless, we were surprised to find that redundancy is still present in the games---particularly those where SER outperformed the baseline.
For example, in Atlantis, we found that nearly 20\% of the replay memory's samples were redundant after 1M training steps, and the most-visited sample was encountered over 250 times.
We believe that SER's performance could be further improved by considering ways to count similar---not just identical---state-action pairs as being redundant (e.g.\ using density models~\cite{ostrovski2017count}).


\paragraph{Conclusion}
SER offers a theoretically well-motivated alternative to the uniform distribution for off-policy deep RL methods.
By correcting for multiplicity bias, SER helps agents learn significantly faster in small MDPs, although the benefits are less pronounced in high-dimensional environments like Atari 2600 games.
We see great promise in future methods that address scalability by exploring ways to generalize over similar state-action pairs during the stratification process.



\begin{acks}
    This work was partially funded by US Army Research Office award W911NF-20-1-0265.
\end{acks}



\clearpage
\bibliographystyle{ACM-Reference-Format} 
\bibliography{references}


\begin{thebibliography}{17}


\ifx \showCODEN    \undefined \def \showCODEN     #1{\unskip}     \fi
\ifx \showDOI      \undefined \def \showDOI       #1{#1}\fi
\ifx \showISBNx    \undefined \def \showISBNx     #1{\unskip}     \fi
\ifx \showISBNxiii \undefined \def \showISBNxiii  #1{\unskip}     \fi
\ifx \showISSN     \undefined \def \showISSN      #1{\unskip}     \fi
\ifx \showLCCN     \undefined \def \showLCCN      #1{\unskip}     \fi
\ifx \shownote     \undefined \def \shownote      #1{#1}          \fi
\ifx \showarticletitle \undefined \def \showarticletitle #1{#1}   \fi
\ifx \showURL      \undefined \def \showURL       {\relax}        \fi
\providecommand\bibfield[2]{#2}
\providecommand\bibinfo[2]{#2}
\providecommand\natexlab[1]{#1}
\providecommand\showeprint[2][]{arXiv:#2}

\bibitem[\protect\citeauthoryear{Bellemare, Naddaf, Veness, and
  Bowling}{Bellemare et~al\mbox{.}}{2013}]%
        {bellemare2013arcade}
\bibfield{author}{\bibinfo{person}{Marc~G Bellemare}, \bibinfo{person}{Yavar
  Naddaf}, \bibinfo{person}{Joel Veness}, {and} \bibinfo{person}{Michael
  Bowling}.} \bibinfo{year}{2013}\natexlab{}.
\newblock \showarticletitle{The Arcade Learning Environment: An Evaluation
  Platform for General Agents}.
\newblock \bibinfo{journal}{\emph{Journal of Artificial Intelligence Research}}
   \bibinfo{volume}{47} (\bibinfo{year}{2013}), \bibinfo{pages}{253--279}.
\newblock


\bibitem[\protect\citeauthoryear{Brockman, Cheung, Pettersson, Schneider,
  Schulman, Tang, and Zaremba}{Brockman et~al\mbox{.}}{2016}]%
        {brockman2016openai}
\bibfield{author}{\bibinfo{person}{Greg Brockman}, \bibinfo{person}{Vicki
  Cheung}, \bibinfo{person}{Ludwig Pettersson}, \bibinfo{person}{Jonas
  Schneider}, \bibinfo{person}{John Schulman}, \bibinfo{person}{Jie Tang},
  {and} \bibinfo{person}{Wojciech Zaremba}.} \bibinfo{year}{2016}\natexlab{}.
\newblock \showarticletitle{Open{AI} Gym}.
\newblock \bibinfo{journal}{\emph{arXiv:1606.01540}} (\bibinfo{year}{2016}).
\newblock


\bibitem[\protect\citeauthoryear{Dietterich}{Dietterich}{2000}]%
        {dietterich2000hierarchical}
\bibfield{author}{\bibinfo{person}{Thomas~G Dietterich}.}
  \bibinfo{year}{2000}\natexlab{}.
\newblock \showarticletitle{Hierarchical Reinforcement Learning with the MAXQ
  Value Function Decomposition}.
\newblock \bibinfo{journal}{\emph{Journal of Artificial Intelligence Research}}
   \bibinfo{volume}{13} (\bibinfo{year}{2000}), \bibinfo{pages}{227--303}.
\newblock


\bibitem[\protect\citeauthoryear{Fedus, Ramachandran, Agarwal, Bengio,
  Larochelle, Rowland, and Dabney}{Fedus et~al\mbox{.}}{2020}]%
        {fedus2020revisiting}
\bibfield{author}{\bibinfo{person}{William Fedus}, \bibinfo{person}{Prajit
  Ramachandran}, \bibinfo{person}{Rishabh Agarwal}, \bibinfo{person}{Yoshua
  Bengio}, \bibinfo{person}{Hugo Larochelle}, \bibinfo{person}{Mark Rowland},
  {and} \bibinfo{person}{Will Dabney}.} \bibinfo{year}{2020}\natexlab{}.
\newblock \showarticletitle{Revisiting Fundamentals of Experience Replay}.
\newblock \bibinfo{journal}{\emph{arXiv:2007.06700}} (\bibinfo{year}{2020}).
\newblock


\bibitem[\protect\citeauthoryear{Fujimoto, Van~Hoof, and Meger}{Fujimoto
  et~al\mbox{.}}{2018}]%
        {fujimoto2018addressing}
\bibfield{author}{\bibinfo{person}{Scott Fujimoto}, \bibinfo{person}{Herke
  Van~Hoof}, {and} \bibinfo{person}{David Meger}.}
  \bibinfo{year}{2018}\natexlab{}.
\newblock \showarticletitle{Addressing Function Approximation Error in
  Actor-Critic Methods}.
\newblock \bibinfo{journal}{\emph{arXiv:1802.09477}} (\bibinfo{year}{2018}).
\newblock


\bibitem[\protect\citeauthoryear{Haarnoja, Zhou, Abbeel, and Levine}{Haarnoja
  et~al\mbox{.}}{2018}]%
        {haarnoja2018soft}
\bibfield{author}{\bibinfo{person}{Tuomas Haarnoja}, \bibinfo{person}{Aurick
  Zhou}, \bibinfo{person}{Pieter Abbeel}, {and} \bibinfo{person}{Sergey
  Levine}.} \bibinfo{year}{2018}\natexlab{}.
\newblock \showarticletitle{Soft Actor-Critic: Off-Policy Maximum Entropy Deep
  Reinforcement Learning with a Stochastic Actor}.
\newblock \bibinfo{journal}{\emph{arXiv:1801.01290}} (\bibinfo{year}{2018}).
\newblock


\bibitem[\protect\citeauthoryear{Kingma and Ba}{Kingma and Ba}{2014}]%
        {kingma2014adam}
\bibfield{author}{\bibinfo{person}{Diederik~P Kingma} {and}
  \bibinfo{person}{Jimmy Ba}.} \bibinfo{year}{2014}\natexlab{}.
\newblock \showarticletitle{Adam: A Method for Stochastic Optimization}.
\newblock \bibinfo{journal}{\emph{arXiv:1412.6980}} (\bibinfo{year}{2014}).
\newblock


\bibitem[\protect\citeauthoryear{Lillicrap, Hunt, Pritzel, Heess, Erez, Tassa,
  Silver, and Wierstra}{Lillicrap et~al\mbox{.}}{2015}]%
        {lillicrap2015continuous}
\bibfield{author}{\bibinfo{person}{Timothy~P Lillicrap},
  \bibinfo{person}{Jonathan~J Hunt}, \bibinfo{person}{Alexander Pritzel},
  \bibinfo{person}{Nicolas Heess}, \bibinfo{person}{Tom Erez},
  \bibinfo{person}{Yuval Tassa}, \bibinfo{person}{David Silver}, {and}
  \bibinfo{person}{Daan Wierstra}.} \bibinfo{year}{2015}\natexlab{}.
\newblock \showarticletitle{Continuous Control with Deep Reinforcement
  Learning}.
\newblock \bibinfo{journal}{\emph{arXiv:1509.02971}} (\bibinfo{year}{2015}).
\newblock


\bibitem[\protect\citeauthoryear{Lin}{Lin}{1992}]%
        {lin1992self}
\bibfield{author}{\bibinfo{person}{Long-Ji Lin}.}
  \bibinfo{year}{1992}\natexlab{}.
\newblock \showarticletitle{Self-Improving Reactive Agents Based on
  Reinforcement Learning, Planning and Teaching}.
\newblock \bibinfo{journal}{\emph{Machine Learning}} \bibinfo{volume}{8},
  \bibinfo{number}{3-4} (\bibinfo{year}{1992}), \bibinfo{pages}{293--321}.
\newblock


\bibitem[\protect\citeauthoryear{Liu and Zou}{Liu and Zou}{2018}]%
        {liu2018effects}
\bibfield{author}{\bibinfo{person}{Ruishan Liu} {and} \bibinfo{person}{James
  Zou}.} \bibinfo{year}{2018}\natexlab{}.
\newblock \showarticletitle{The Effects of Memory Replay in Reinforcement
  Learning}. In \bibinfo{booktitle}{\emph{Allerton Conference on Communication,
  Control, and Computing}}. IEEE, \bibinfo{pages}{478--485}.
\newblock


\bibitem[\protect\citeauthoryear{Mnih, Kavukcuoglu, Silver, Rusu, Veness,
  Bellemare, Graves, Riedmiller, Fidjeland, Ostrovski, et~al\mbox{.}}{Mnih
  et~al\mbox{.}}{2015}]%
        {mnih2015human}
\bibfield{author}{\bibinfo{person}{Volodymyr Mnih}, \bibinfo{person}{Koray
  Kavukcuoglu}, \bibinfo{person}{David Silver}, \bibinfo{person}{Andrei~A
  Rusu}, \bibinfo{person}{Joel Veness}, \bibinfo{person}{Marc~G Bellemare},
  \bibinfo{person}{Alex Graves}, \bibinfo{person}{Martin Riedmiller},
  \bibinfo{person}{Andreas~K Fidjeland}, \bibinfo{person}{Georg Ostrovski},
  {et~al\mbox{.}}} \bibinfo{year}{2015}\natexlab{}.
\newblock \showarticletitle{Human-Level Control through Deep Reinforcement
  Learning}.
\newblock \bibinfo{journal}{\emph{Nature}} \bibinfo{volume}{518},
  \bibinfo{number}{7540} (\bibinfo{year}{2015}), \bibinfo{pages}{529--533}.
\newblock


\bibitem[\protect\citeauthoryear{Ostrovski, Bellemare, Oord, and
  Munos}{Ostrovski et~al\mbox{.}}{2017}]%
        {ostrovski2017count}
\bibfield{author}{\bibinfo{person}{Georg Ostrovski}, \bibinfo{person}{Marc~G
  Bellemare}, \bibinfo{person}{A{\"a}ron Oord}, {and} \bibinfo{person}{R{\'e}mi
  Munos}.} \bibinfo{year}{2017}\natexlab{}.
\newblock \showarticletitle{Count-Based Exploration with Neural Density
  Models}. In \bibinfo{booktitle}{\emph{International Conference on Machine
  Learning}}. PMLR, \bibinfo{pages}{2721--2730}.
\newblock


\bibitem[\protect\citeauthoryear{Sharma, Pal, Anand, and Kaul}{Sharma
  et~al\mbox{.}}{2020}]%
        {sharma2020stratified}
\bibfield{author}{\bibinfo{person}{Anil Sharma}, \bibinfo{person}{Mayank~K
  Pal}, \bibinfo{person}{Saket Anand}, {and} \bibinfo{person}{Sanjit~K Kaul}.}
  \bibinfo{year}{2020}\natexlab{}.
\newblock \showarticletitle{Stratified Sampling Based Experience Replay for
  Efficient Camera Selection Decisions}. In \bibinfo{booktitle}{\emph{IEEE
  International Conference on Multimedia Big Data}}. IEEE,
  \bibinfo{pages}{144--151}.
\newblock


\bibitem[\protect\citeauthoryear{Sutton and Barto}{Sutton and Barto}{2018}]%
        {sutton2018reinforcement}
\bibfield{author}{\bibinfo{person}{Richard~S Sutton} {and}
  \bibinfo{person}{Andrew~G Barto}.} \bibinfo{year}{2018}\natexlab{}.
\newblock \bibinfo{booktitle}{\emph{Reinforcement Learning: An Introduction}}.
\newblock \bibinfo{publisher}{MIT Press}.
\newblock


\bibitem[\protect\citeauthoryear{Wang, Bapst, Heess, Mnih, Munos, Kavukcuoglu,
  and de~Freitas}{Wang et~al\mbox{.}}{2016}]%
        {wang2016sample}
\bibfield{author}{\bibinfo{person}{Ziyu Wang}, \bibinfo{person}{Victor Bapst},
  \bibinfo{person}{Nicolas Heess}, \bibinfo{person}{Volodymyr Mnih},
  \bibinfo{person}{Remi Munos}, \bibinfo{person}{Koray Kavukcuoglu}, {and}
  \bibinfo{person}{Nando de Freitas}.} \bibinfo{year}{2016}\natexlab{}.
\newblock \showarticletitle{Sample Efficient Actor-Critic with Experience
  Replay}.
\newblock \bibinfo{journal}{\emph{arXiv:1611.01224}} (\bibinfo{year}{2016}).
\newblock


\bibitem[\protect\citeauthoryear{Watkins}{Watkins}{1989}]%
        {watkins1989learning}
\bibfield{author}{\bibinfo{person}{Christopher John Cornish~Hellaby Watkins}.}
  \bibinfo{year}{1989}\natexlab{}.
\newblock \emph{\bibinfo{title}{Learning from Delayed Rewards}}.
\newblock \bibinfo{thesistype}{Ph.D. Dissertation}. \bibinfo{school}{King's
  College}.
\newblock


\bibitem[\protect\citeauthoryear{Zhang and Sutton}{Zhang and Sutton}{2017}]%
        {zhang2017deeper}
\bibfield{author}{\bibinfo{person}{Shangtong Zhang} {and}
  \bibinfo{person}{Richard~S Sutton}.} \bibinfo{year}{2017}\natexlab{}.
\newblock \showarticletitle{A Deeper Look at Experience Replay}.
\newblock \bibinfo{journal}{\emph{arXiv:1712.01275}} (\bibinfo{year}{2017}).
\newblock


\end{thebibliography}


\end{document}